# ReckOn: A 28-nm Sub-mm² Task-Agnostic Spiking Recurrent Neural Network Processor Enabling On-Chip Learning over Second-Long Timescales

Charlotte Frenkel, Giacomo Indiveri

University of Zurich and ETH Zurich, Zurich, Switzerland

The robustness of autonomous inference-only devices deployed in the real world is limited by data distribution changes induced by different users, environments, and task requirements. This challenge calls for the development of edge devices with an always-on adaptation to their target ecosystems. However, the memory requirements of conventional neural network training algorithms scale with the temporal depth of the data being processed, which is not compatible with the constrained power and area budgets at the edge. For this reason, previous works demonstrating end-to-end on-chip learning without external memory were restricted to the processing of static data such as images [1-4], or to instantaneous decisions involving no memory of the past, e.g. obstacle avoidance in mobile robots [5]. The ability to learn short- to long-term temporal dependencies on-chip is a missing enabler for robust autonomous edge devices in applications such as gesture recognition, speech processing, and cognitive robotics.

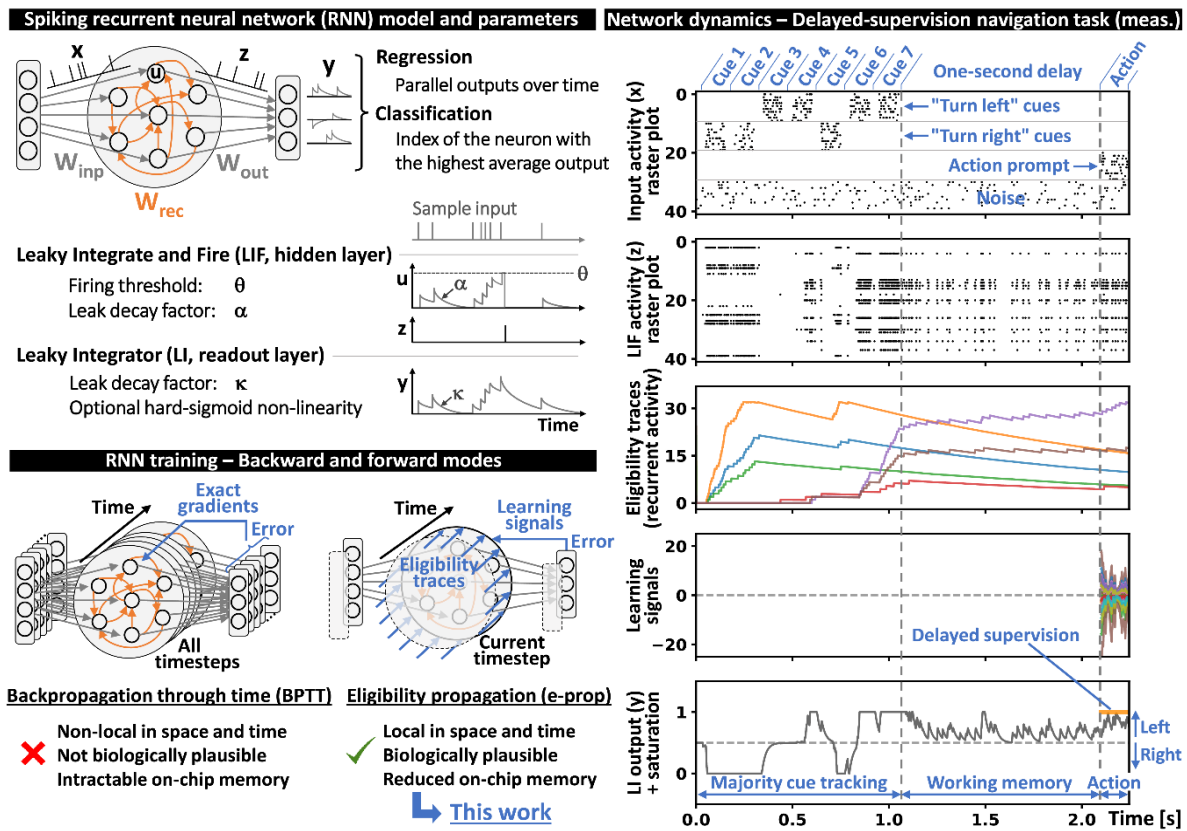

Fig. 1: Key concepts for an on-chip online training of spiking RNNs over long timescales (left). Delayed-supervision navigation task requiring temporal credit assignment over seconds (right).

To solve this challenge, we follow a neuromorphic approach and introduce RecKon, a spiking recurrent neural network (RNN) processor enabling online learning over seconds with a millisecond temporal resolution. It exploits (1) feed-forward eligibility traces (ETs) for a bio-inspired approximation to the backpropagation through time (BPTT) algorithm that is local in both space and time (Fig. 1, left), thereby drastically reducing the memory requirements, (2) sparsity in the input data and weight updates to minimize the computational footprint of on-demand processing, and (3) the address-event (AE) representation used by spiking retina and cochlea sensors for task-agnostic processing. These key features allow demonstrating end-to-end on-chip training on temporal data within sub-150μW and sub-mm² power and area budgets for tasks spanning gesture recognition, keyword spotting (KWS), and navigation, accelerated up to 98× at 0.5V.

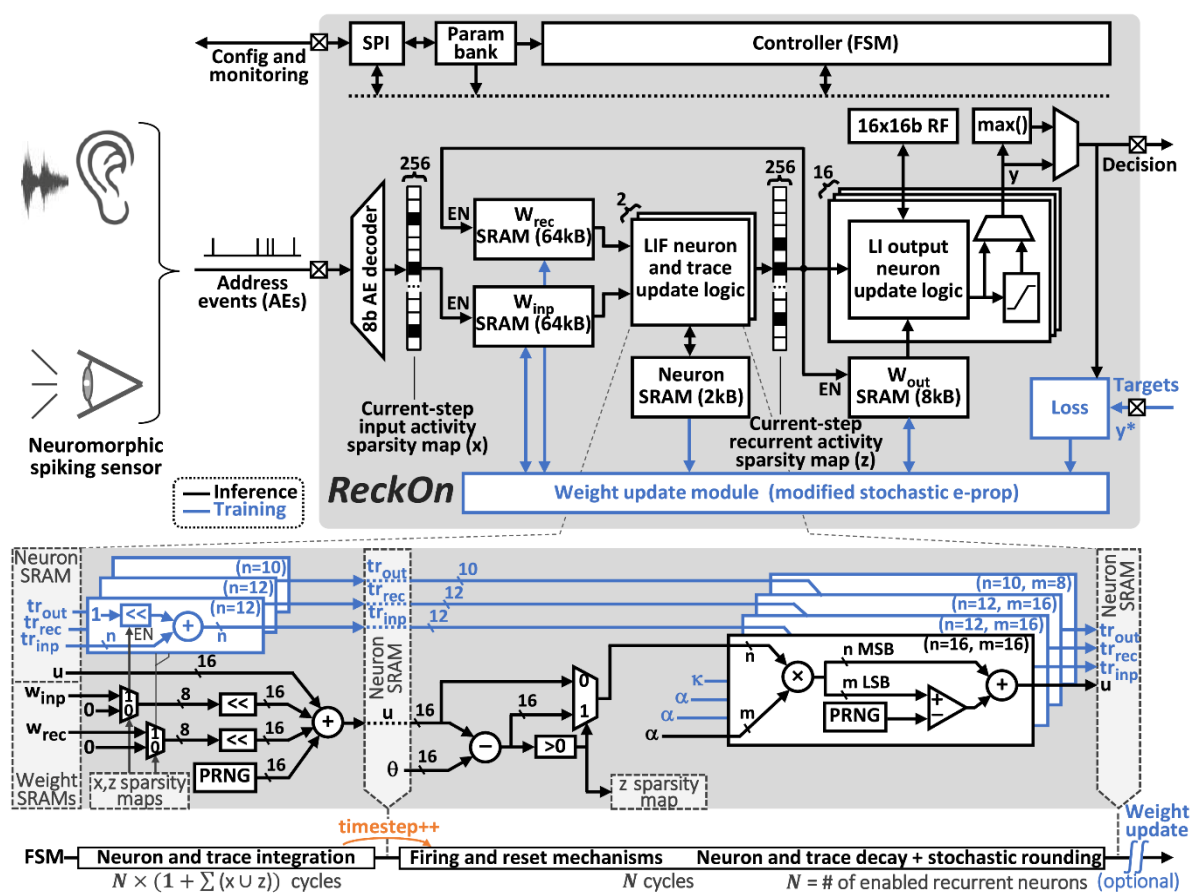

Fig. 2: System diagram of ReckOn (top). Architecture of the LIF neuron and trace update logic with forward-pass control flow at each timestep (bottom).

Figure 2 shows the system diagram and the architecture of ReckOn, which implements the spiking RNN topology outlined in Fig. 1 (top left) and computes the network dynamics in a time-stepped fashion. Input spike events produced by asynchronous neuromorphic sensors are handled through an 8-bit 4-phase-handshake AE decoder. The recurrent hidden layer consists of up to 256 leaky integrate-and-fire (LIF) neurons with all-to-all input and recurrent connectivity through 8-bit weights stored in two 64-kB SRAMs. The neuron states are stored in a 2-kB SRAM, where each 128-bit word stores the states of two LIF neurons and their shared leakage and threshold parameters. Input and recurrent spiking activities are buffered into binary sparsity maps until they are processed at the next timestep. For every non-zero entry in the sparsity maps, the corresponding synaptic weights are read and two time-multiplexed instances of the LIF update logic proceed by (1) adding the weighted input and recurrent activities to the membrane potential, (2) assessing the threshold crossing condition and the associated reset-by-subtraction mechanism, and (3) computing the membrane decay corresponding to the neuron leakage. Finally, the output layer consists of up to 16 leaky-integrator (LI) neurons, which are processed identically to LIF neurons, except for the absence of the spike and reset mechanisms. An optional hard-sigmoid activation can be applied. This real-valued readout avoids resorting to ill-defined loss functions relying on a distance metric between two spike trains. Both regression and classification tasks can be handled.

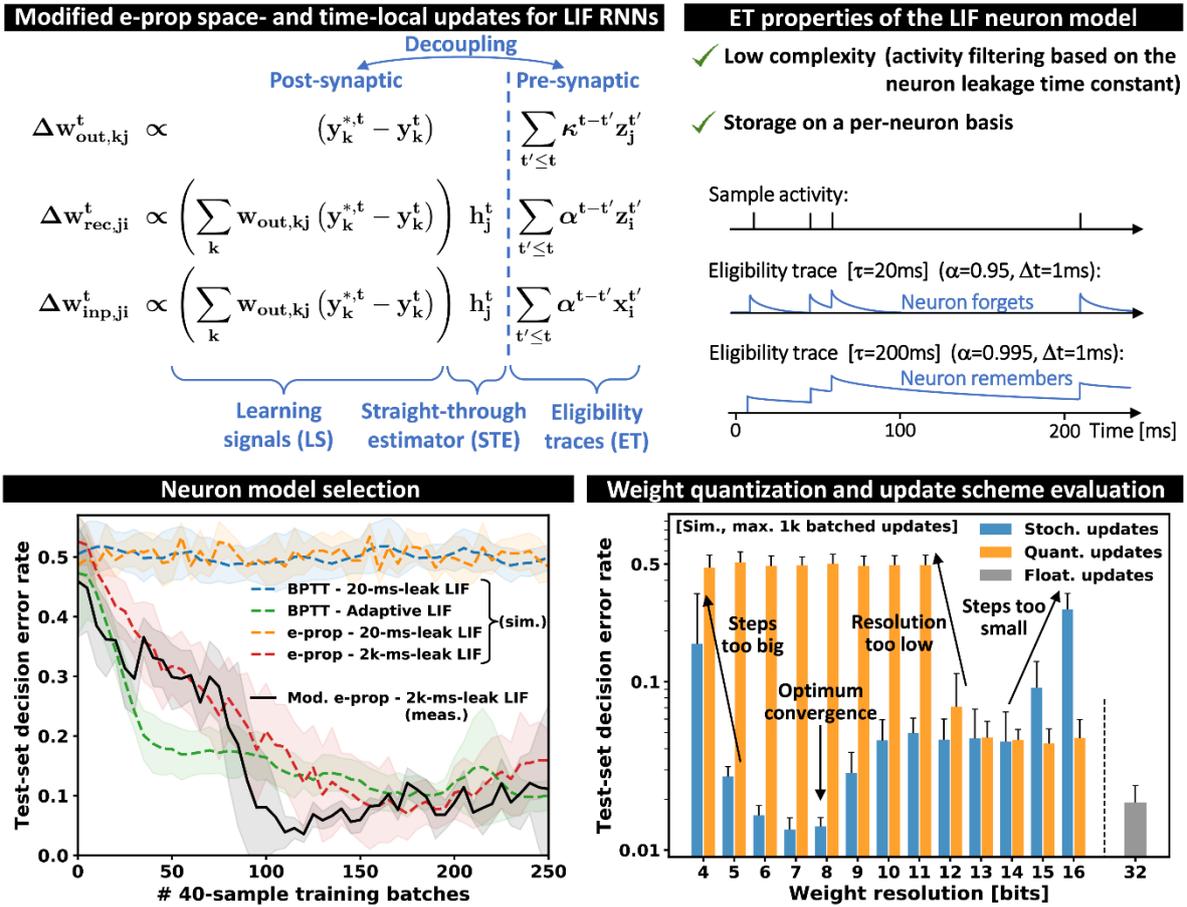

Fig. 3: Modifications to e-prop and evaluation on the navigation task (symbols as defined in Fig. 29.4.1, experiments averaged over 10 trials).

By approximating BPTT as a product of error-dependent learning signals and ETs computed in the feed-forward path (Fig. 1, bottom left), the eligibility propagation (e-prop) algorithm proposed in [6] sidesteps the need to unroll the network in time during training, an issue whose memory penalty often requires truncating BPTT to short temporal depths. It is also shown in [6] that networks of LIF neurons with a bio-plausible 20-ms leakage time constant have insufficient dynamics to learn long temporal dependencies, as opposed to networks of adaptive

LIF (ALIF) neurons, which embed firing threshold adaptation over hundreds of milliseconds. However, as the computation of ETs in ALIF-based e-prop requires a complex multi-timescale filtering in each synapse of the network, its computational and memory requirements are not amenable to an on-chip implementation. We address this challenge in three steps, which we illustrate for the delayed-supervision navigation task described in Fig. 1 (right). (1) *Neuron model selection*: we show in Fig. 3 (bottom left) that increasing the leakage time constant of LIF neurons to match the temporal contents of the target task leads to a learning performance comparable to that of ALIF-based RNNs. (2) *Space and time locality*: we simplify the e-prop equations for the LIF neuron model as per Fig. 3 (top). The resulting weight updates are applied on a per-timestep basis and consist of three factors available locally at the synapse: the ET, the learning signal (LS) and the straight-through estimator (STE) of the spiking activation function derivative. For each of the input, recurrent, and output weight updates, there is a complete decoupling between the pre-synaptic (ET) and post-synaptic terms (LS and STE). The computation and storage of ETs thus scales with the number of neurons, instead of the number of synapses. (3) *Weight resolution and update scheme*: we show in Fig. 3 (bottom right) that applying updates stochastically to 8-bit synaptic weights results in a fast training convergence at no extra weight memory footprint compared to the feed-forward phase. Therefore, without degrading the network learning performance, the combination of these techniques reduces the memory overhead for on-chip learning over long timescales to only 0.8% of the equivalent inference-only design. The memory overhead solely consists of the added per-neuron storage of ETs.

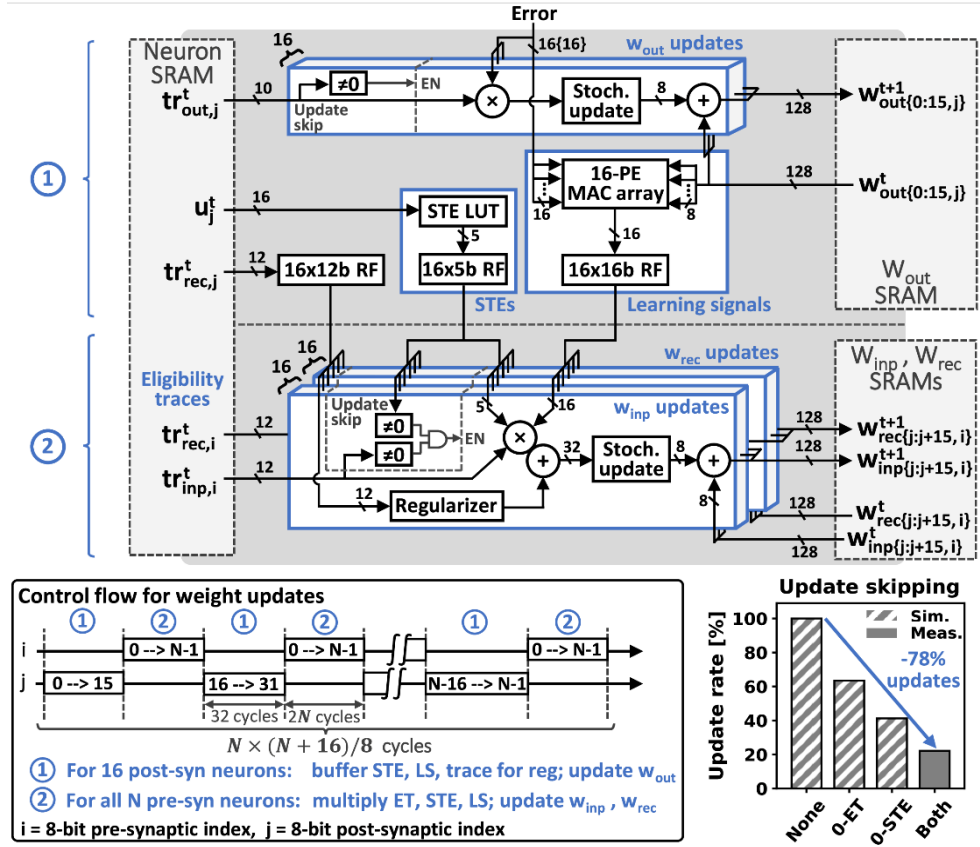

Fig. 4: Weight update circuit architecture (top) and control flow (bottom left). Sparsity can be exploited for weight updates with ET- and STE-based skipping, shown for the navigation task (bottom right).

Figure 4 shows the circuit implementation of stochastic weight updates based on the modified e-prop LIF equations. A 16-way time-multiplexing approach meets real-time processing constraints while matching the 128-bit word length of the weight SRAMs, whose aspect ratio was chosen to maximize density. The control flow diagram of Fig. 4 (bottom) consists of two alternating steps: (1) the output weights are updated and the associated post-synaptic STE and LS terms are buffered; (2) the input and recurrent weight updates are carried out by multiplying the buffered post-synaptic STE and LS terms with the pre-synaptic ETs, whose computation took place in the feed-forward phase (Fig. 2, bottom). As they provide a low-pass-filtered image of the network activity, ETs are also conveniently used for weight regularization. The STE is implemented as a programmable 5-segment 5-bit signed function in a local LUT. While sparsity is leveraged in the feed-forward phase by only processing the non-zero elements of the input- and recurrent-activity sparsity maps, the sparsity of the STE function and of inactive-neuron ETs can be exploited in the weight update phase. In Fig. 4 (bottom right), we show that combining ET- and STE-based skipping reduces the weight update rate by 78%.

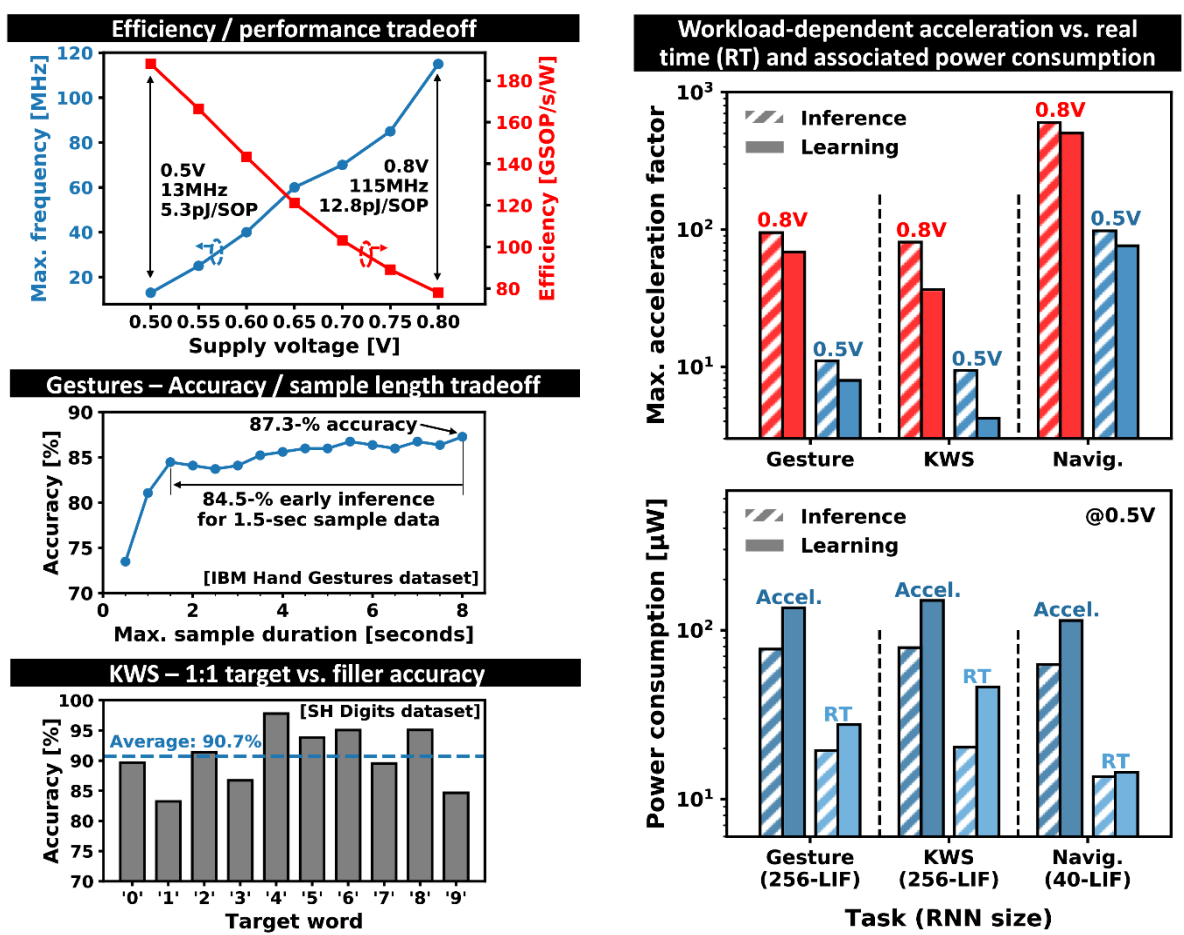

Fig. 5: Benchmarking and measurement results (minimum operating frequency and average energy efficiency for 5 chips in 28nm CMOS at 22°C).

|  | This work | VLSI'15 [1] | VLSI'17 [7] | ISSCC'18 [2] | VLSI'18 [3] | ISSCC'19 [4] | ISSCC'18 [5] |
|---|---|---|---|---|---|---|---|
| Technology | 28nm | 65nm | 40nm | 65nm | 10nm | 65nm | 55nm |
| Implementation | Digital | Digital | Mixed-signal | Mixed-signal (IMC) | Digital | Digital | Mixed-signal |
| Core area | 0.45mm² | 1.8mm² | 1.3mm² | 0.8mm² | 1.72mm² | 10.1mm² | 3.1mm² |
| Memory | 138kB | 37.6kB | N/A | 16kB | 896kB | 353kB | 0.4kB |
| Energy metric | 5.3pJ/SOP [a] | 5.7pJ/pix | 48.9pJ/pix | 0.32pJ/OP | 3.8pJ/SOP | 0.29pJ/OP | 0.32pJ/OP |
| Network type | Spiking RNN | Spiking LCA [b] | Spiking LCA [b] | SVM | Multicore SNN | Binary NN | ANN |
| # Neurons | (256)-256-16 | 4x64 | 8x64 | 128 | 64x64 | (784)-200-200-10 | (3) - 84 - 3 |
| # Synapses (width) | 132k (8-bit) | 83k (4,5,14-bit) | N/A | 8k (16-bit) | 1M (7-bit) | 194k (14-bit) | 0.5k (6-bit) |
| On-chip learning | ✓ | ✓ | ✓ | ✓ | ✓ | ✓ | ✓ |
| - algorithm | Mod. stoch. e-prop | SGD | N/A | SGD | STDP | Mod. SD [c] | SGD |
| - multilayer | ✓ | ✗ | ✗ | ✗ | ✗ | ✓ | ✓ |
| - dynamics | Few ms to seconds | ✗ | ✗ | ✗ | Few ms | ✗ | ✗ |
| Task | Hand gesture classif. Keyword spotting Navigation | Image classif. | Image classif. | Image classif. | Image classif. | Image classif. | Obstacle avoid. |
| Dataset | IBM DVS Gestures [f] Spiking Heidelberg Digits [g] Delayed cue integration [h] | MNIST | MNIST | MIT CBCL [d] | MNIST [e] | MNIST | Custom autonomous robot |
| Average input data depth | Gest: 1318 steps @Δt=5ms KWS: 104 steps @Δt=5ms Nav: 2250 steps @Δt=1ms | 1 frame | 1 frame | 1 frame | 1 frame | 1 frame | N/A (1-step decisions) |
| Accuracy with on-chip training | Gest: 87.3% @10classes KWS: 90.7% @1word Nav: 96.4% @2decisions | 84%-90% | 88% | 91.6% | 89% | 97.8% | N/A |
| Power (infer / learn) | Gest: 77µW / 135µW [a] KWS: 79µW / 150µW [a] Nav: 62µW / 114µW [a] | 268mW / 526mW | 87mW / N/A | 1.3mW / 3.1mW | 6.2mW / N/A | 23.6mW / 23.1mW | 690µW / N/A |
| Energy per step (infer / learn) | Gest: 35nJ / 85nJ [a] KWS: 42nJ / 178nJ [a] Nav: 0.6nJ / 1.5nJ [a] | 27-162nJ / 94.7µJ | 50.1nJ / N/A | 42pJ / 150pJ | 1.0µJ / N/A | 236nJ / 254nJ | 0.69nJ / 1.5nJ |

[a] At 0.5V, 13MHz, accelerated-time  [b] Locally-competitive algorithm  [c] Segregated dendrites algorithm  [d] Downscaled to 11x11  [e] Pre-processed with Gabor filters
[f] From [8], downscaled to 16x16, 10 classes   [g] From [9], English digits 0-9, channel subsampling 1:3, target vs. filler word ratio 1:1   [h] As specified in [6]

Fig. 6: Comparison with prior works embedding end-to-end on-chip learning (i.e. full on-chip weight storage, no off-chip pre-training, accuracy results from on-chip training only).

With all weights randomly initialized, we show sensor- and task-agnostic end-to-end on-chip learning with three benchmarks that incorporate second-long timescales (Figs. 5-6): (1) with the spiking-retina IBM DVS Gestures dataset, we demonstrate 87.3-%-accuracy 10-class hand gesture classification with a flexible latency-accuracy tradeoff (Fig. 5, middle left); (2) with the spiking-cochlea Spiking Heidelberg Digits dataset, we demonstrate 90.7-%-accuracy 1-word KWS in a 1:1 target versus filler word setup (Fig. 5, bottom left); (3) with synthetic data typical of behavioral-timescale learning analysis in rodents (Fig. 1, right), we demonstrate 96.4-%-accuracy binary-decision navigation. The latter benchmark also outlines direct applications in delayed-reward reinforcement learning and robotic simultaneous localization and mapping (SLAM). For all tasks, the power consumption and the maximum acceleration compared to a real-time processing of incoming data depend on the workload (sparsity, network resources, and training time). Figure 5 (right) shows that the nominal supply voltage of 0.8V at a clock frequency of 115MHz allows for an accelerated-time processing of pre-generated datasets by 37× to 600×. The supply voltage can be scaled down to 0.5V, which allows for (1) accelerated processing by 4× to 98× within power budgets of 150µW (learning) and 80µW (inference) at 13MHz, or (2) real-time processing of spikes generated on-the-fly by neuromorphic sensors within 46µW (learning) and 20µW (inference). Fig. 5 (top left) shows a peak efficiency of 5.3pJ per synaptic operation (SOP) at 0.5V.

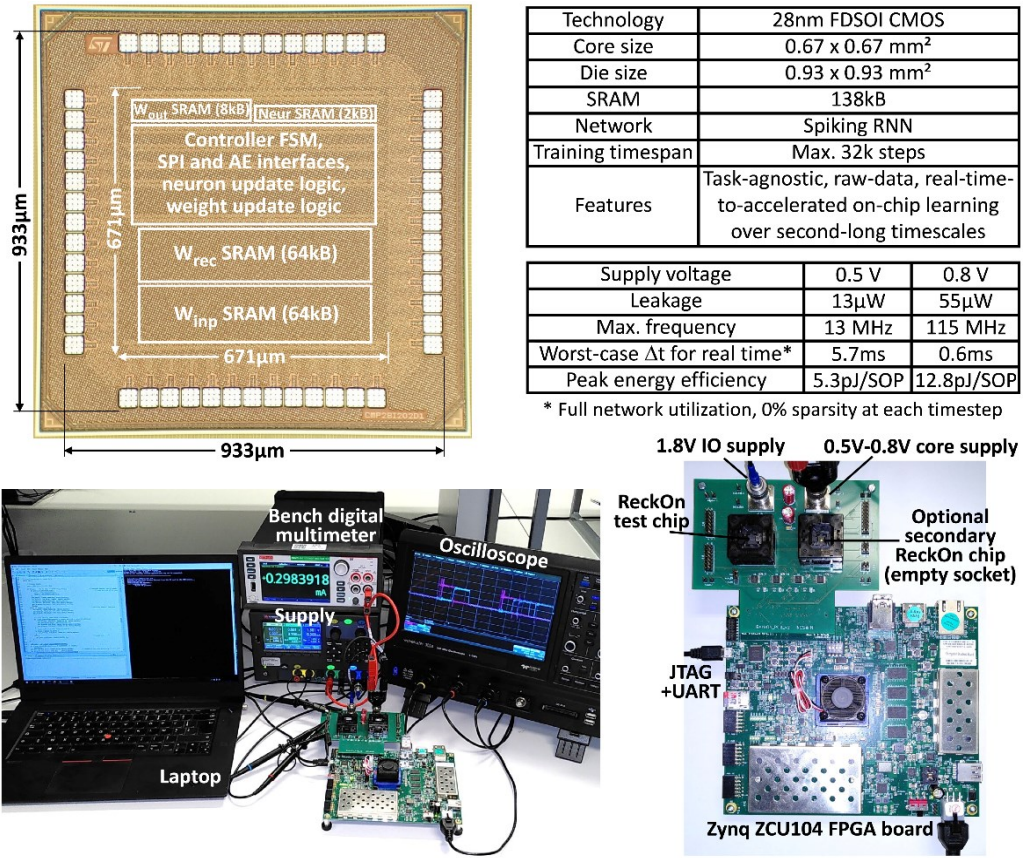

Fig. 7: Die microphotograph, chip summary and measurement setup.

Figure 7 shows the 28-nm FDSOI CMOS die microphotograph. ReckOn has a core area of only 0.45mm² with 138kB of SRAM storage. It is compared in Fig. 6 with prior works embedding end-to-end on-chip learning, without external memory nor pre-training, all of which carry out task-specific learning and inference over a single frame or timestep. With competitive energy per step values of 0.6-42nJ (inference) and 1.5-178nJ (learning) at 0.5V, the task-agnostic ability of ReckOn to learn over thousands of timesteps is unique and opens the doors to chip repurposing and sensor fusion in autonomous edge devices. ReckOn is an open-source design available at https://github.com/ChFrenkel/ReckOn.

*Acknowledgments:*

Funding by the EU and SNSF (No. 826655, 20CH21186999/1). Circuit fabrication by ST.